\documentclass{article}

\usepackage[preprint]{nips_2018}

\usepackage[utf8]{inputenc} 
\usepackage[T1]{fontenc}    
\usepackage{hyperref}       
\usepackage{url}            
\usepackage{booktabs}       
\usepackage{amsfonts}       
\usepackage{nicefrac}       
\usepackage{microtype}      
\usepackage[pdftex]{graphicx}
\usepackage{subfigure}
\usepackage{caption}
\usepackage{multirow}

\title{Learning multiple non-mutually-exclusive tasks for improved classification of inherently ordered labels}

\author{
  Vadim Ratner, Yoel Shoshan, Tal Kachman\\
  IBM Research\\
  Haifa Research Labs\\
  Haifa, Israel 3498825 \\
  vadimra@il.ibm.com, yoels@il.ibm.com, tal.kachman@ibm.com \\
}

\begin{document}

\maketitle

\begin{abstract}
  Medical image classification involves thresholding of labels that represent malignancy risk levels. Usually, a task defines a single threshold, and when developing computer-aided diagnosis tools, a single network is trained per such threshold, e.g. as screening out healthy (very low risk) patients to leave possibly sick ones for further analysis (low threshold), or trying to find malignant cases among those marked as non-risk by the radiologist ("second reading", high threshold). We propose a way to rephrase the classification problem in a manner that yields several problems (corresponding to different thresholds) to be solved simultaneously. This allows the use of Multiple Task Learning (MTL) methods, significantly improving the performance of the original classifier, by facilitating effective extraction of information from existing data. 
\end{abstract}

\section{Introduction}

Recent progress in the Deep Learning field has brought many natural image classification tasks to better than human performance (e.g. [5]). While performance in the medical field falls somewhat behind due to general sparsity of labeled data, it too has greatly benefited from Deep-Learning-based algorithms [4].

The most common approach in classification of medical images is adjusting networks developed for natural images, with the main challenge being the collection of data. However, we believe that medical tasks have some unique characteristics that are worth exploiting. Resulting algorithms can also be utilized outside of the medical domain.

One such difference is the inherent order within most medical labels. A diagnosis of a disease is a risk estimation problem, and while the patient is usually interested in a binary result (healthy or sick), the medical practitioner prefers a finer division of labels that represent levels of risk of malignancy. The estimated risk level then dictates further course of treatment.

An example of such label system is BIRADS: Breast Imaging-Reporting and Data System. BIRADS include five diagnostic labels – ‘1’ – an image only contains healthy tissue, ‘2’ – an image contains only benign findings, ‘3’ – probably benign (less than 2\% risk), ‘4’ – an image may contain malignant findings, and ‘5’ – an image probably contains malignant findings. The BIRADS system also has two special labels - '0' meaning that further analysis is required (usually more imaging), and '6' - cases with existing biopsy results. In this case the ordering of the diagnostic labels is obvious – higher numbers mean higher probability of malignancy. BIRADS 3 and above require follow-up, with BIRADS 4 and above demanding biopsy (and in some cases immediate surgery). Many more similar systems are being developed, such as LIRADS (liver), GIRADS (gynecology), CRADS (colonography), and more.

Machine learning tasks for such cases often involve making a binary decision regarding a suggested treatment, based on some thresholding of the label range. The threshold choice is often task-related, and there may be several viable thresholds. In the case of screening, it may be interesting to distinguish between BIRADS 1 (benign), and 2-5, or 1-2 vs. 3-5 with the higher risk group requiring further analysis by a radiologist, dismissing some of the obviously healthy patients, thus saving radiologists’ time. Discriminating between BIRADS 1-3 and 4-5 is useful in reviewing radiologist decisions to avoid false negatives.

Three problems tend to come up when solving such tasks: 
\begin{itemize}
\item The labels by themselves contain no information about the order between them. This information must be conveyed to the learning algorithm separately.
\item The distinction between adjacent labels is often unclear and annotator-dependent, leading to a situation where the same input may be marked with different (although probably adjacent) labels by different practitioners (or even by the same one). Radiologist usually agree on malignancy of less than 80\% cases ([8]), whereas agreement rate may be even lower on predicting individual BIRADS labels ([10]).
\item Sometimes not all threshold values allow efficient training because they result in highly unbalanced data sets. Especially in medicine, higher thresholds, such as BIRADS 1-3 vs 4-5, result in large negative and very small positive groups (actual cancer occurs in only about 0.5\% of the population), while lower thresholds, such as BIRADS 1 vs 2-5, may result in balanced positive/negative groups.
\end{itemize}

We propose to alleviate these problems by simultaneous training on additional tasks defined by groupings labels differently than for the original task.

\section{Previous work}
\label{previous}

The most commonly used approach to medical classification problems is to define a single division of data. The division is usually binary - malignant vs. benign (e.g. [4]), however, some authors divide the risk spectrum into several classes, such as defined by Geras et al. ([3]), who classified mammogram images into healthy (BIRADS1), benign (BIRADS 2), and requiring further analysis (BIRADS 0, containing, in this case, BIRADS 3, 4, and 5). The resulting classes are non-overlapping, and a single network is trained for each task. This approach originates from natural image classification, where the classes are usually well-defined (there is no smooth transition between a cat and a dog, for example). In the case of ordinal labels, however, while some are always positive (or negative), such as ones corresponding to lowest (and highest) risk levels, others may belong to either one of the groups (mid-level risk may be considered either risky or safe), depending on the problem at hand. In this paper we plan to exploit this property.

In some cases, additional data is utilized in order to exploit the strength of Multiple Task Learning (MTL). In this setting, the network is trained on several related tasks, such as classification by malignancy and by visual characteristics, with the representation being shared between the tasks ([6], [7]), helping the network to generalize better. 

There is a general approach of finding inter-label relations that usually involves adding layers to the network (e.g. [9]), however this requires more labeled data, which is difficult to obtain in the medical domain. 

A somewhat similar setting to the one proposed in this paper, only with the opposite problem, is encountered in the field of learning from aggregate views [2]. There the labels have been aggregated in different ways, and one needs to train a classifier without having access to the non-aggregated labels (due to encryption, or privacy concerns).

Another closely related field if curriculum learning [1]. There, the network is trained on progressively more complex tasks, reducing training time and improving overall performance.


\section{Proposed method}
\label{method}

Let us consider a problem of training a binary classifier for a given task, with the ground-truth generated by thresholding a group of ordinal labels, say, risk levels from 1 to 5, with a threshold \emph{T}. We propose a type of Multi-Task Learning algorithm (MTL) that does not require any extra data (that may not always be available), and where the extra tasks are classifiers based on additional thresholds. This way, the positive groups of labels for the different tasks will become progressively more "cautious" as the task threshold decreases (i.e. consider lower label values as "risky").
 
Positive groups of labels created in this manner overlap, thus achieving 2 goals:
\begin{itemize}
	\item By grouping the labels by degrees of caution we provide information about relations between the labels.
	\item Lowering the impact of label noise. Now, a shift of a label from value ‘a’ to its neighboring value ‘b’ only affects one output (the one where the threshold is between the two values) out of several, but not all of them. This way the network is still able to learn from mislabeled data.
\end{itemize}

In addition, the division of data into classes as dictated by the original task is often unbalanced, reducing training performance. There usually exist divisions that result in more balanced sets and these can be used to define additional tasks when training the network. As in all multi-task schemes, training on a related but different (balanced) task improves performance of the desired (unbalanced) one.

Also, training on tasks corresponding to some of the thresholds may be easier. The reasons for this may be data-related (e.g. better balancing, or lower label noise), or task-related (e.g. distinguishing between completely healthy tissue and some finding is usually easier than telling if a finding is malignant or benign). In this respect, training on multiple thresholds simultaneously resembles curriculum learning ([1]), where simpler tasks help “prepare” the network for training on more difficult ones.

\begin{figure}[h]
	\centering
	\includegraphics[width=1.0\linewidth]{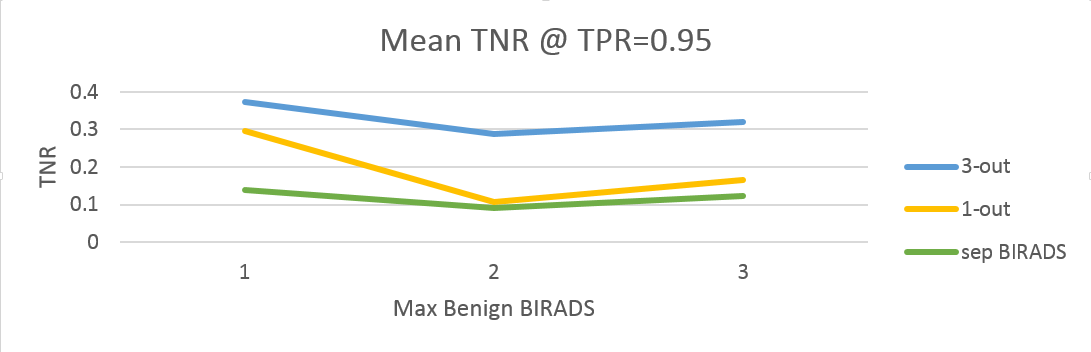}
	\caption{Mean TNR (y-axis) at TPR of 0.95, for 3 tasks (x-axis): BIRADS 1 vs 2-5 (‘1’), 1-2 vs 3-5 (‘2’) and 1-3 vs 4-5 (‘3’). Three architectures are presented: single output per task (yellow), 3 outputs (MTL version, one per task) (blue) and 4 outputs (one per BIRADS value 1,2,3,4-5) which are used to infer the task labels (green). Otherwise the architectures, parameters and train/test sets are the same.}
	\label{fig:results}
\end{figure}

As a result, the proposed method significantly improves results when compared to single output and to direct label prediction (fig. \ref{fig:results})

Any network that learns to distinguish between groups of ordinal labels (either per-pixel for segmentation, or per-image for classification) can be augmented with additional outputs representing different thresholds or divisions between the groups. An example is fully connected final layer with a single output, say, BIRADS 1-2 vs 3-5, will now have several: 1 vs. 2-5, 1-2 vs. 3-5, and 1-3 vs. 4-5. 

The outputs are no longer ‘1-hot’, i.e. ground truth of the resulting output vector may have more than a single ‘1’ value – this should be taken into consideration when choosing a loss function.

\section{Experiments}

\subsection{Setup}

\begin{table}
	\caption{Data distribution between\\ different BIRADS values}
	\label{data}
	\centering
	\begin{tabular}{ll}
		\toprule		
		BIRADS value & US images\\
		\midrule
		1 & 2700 \\
		2 & 1113 \\
		3 & 359 \\
		4-5 & 732 \\
		Total  & 4904 \\
		\bottomrule
	\end{tabular}	
\end{table}

\paragraph{Data:}

We use a dataset of Ultrasound (US) images that consists of 4904 weakly labeled images (i.e. the only data about the image is its BIRADS value)The data distribution is summarized in table \ref{data}. All networks are cross-validated on 5 folds of the data, resulting in 80\% of the data being used for training and 20\% for validation on each fold. The folds are balanced by BIRADS values in such a way that all images belonging to a given patient appear in a single fold, to avoid contamination.

\paragraph{Performance measure:}

We compare True Negative Rates (TNR) at the operation point of True Positive Rate (TPR) of 0.95. This point was chosen because most medical applications require high TPR, in order not to misdiagnose diseased tissue as healthy, and because the amount of data in the validation sets does not allow us to reliably claim higher values of TPR (see, e.g., [11] for further analysis).

\begin{table}
	\caption{Network architecture}
	\label{tabarch}
	\centering
	\begin{tabular}{|c|}
		
		\hline	
		Input \\ \hline
		Conv. layer, 80 filters\\ \hline
		Dropout layer (50\%)\\ \hline
		Conv. layer, 80 filters\\ \hline
		Conv. layer, 80 filters\\ \hline
		Batch Normalization\\ \hline
		Max Pooling (2x2)\\ \hline
		Dropout layer (50\%)\\ \hline
		Conv. layer, 160 filters\\ \hline
		Conv. layer, 160 filters\\ \hline
		Batch Normalization\\ \hline
		Max Pooling (2x2)\\ \hline
		Dropout layer (50\%)\\ \hline
		Conv. layer, 320 filters\\ \hline
		Batch Normalization\\ \hline
		Max Pooling (2x2)\\ \hline
		Dropout layer (50\%)\\ \hline
		Dense Layer (320 outputs)\\ \hline
		1-4 outputs (sigmoid activation)\\ \hline		
		
	\end{tabular}	
\end{table}

\paragraph{Architecture:}

Architecture of the network used is summarized in table \ref{tabarch}. The loss function we use is binary cross-entropy for every output, averaging it over multiple outputs for the MTL version.

\paragraph{Training:}
We train 5 types of classifiers:
\begin{itemize}
	\item Single-output, BIRADS 1 (55\% of the data) vs. 2-5 (45\% of the data), corresponding to BIRADS threshold value of 1.
	\item Single-output, BIRADS 1-2 (78\%) vs. 3-5 (22\%), threshold value of 2.
	\item Single-output, BIRADS 1-3 (85\%) vs. 4-5 (15\%), threshold value of 3.
	\item Proposed multi-task classifier with three outputs: BIRADS 1 vs. 2-5; 1-2 vs. 3-5; and 1-3 vs. 4-5 (corresponding to thresholds of 1, 2, and 3), with every output performance compared separately to that of a single-output classifier trained for the same task.
	\item A BIRADS value classifier, with one-hot output vector that has four states: BIRADS 1 (55\% of the data), BIRADS 2 (23\%), BIRADS 3 (7\%), and BIRADS  4-5 (15\%). The output is then used to derive the BIRADS value, which thresholded and compared to the outputs of the other networks.	 
\end{itemize}

\subsection{Results}

The results are depicted in fig. \ref{fig:results}. As expected, the task of distinguishing between BIRADS 1 and BIRADS 2-5 seems to be the easiest for all the tested networks, with single-output architecture and the proposed method achieving similar performance (slightly higher for the proposed method). The tasks of BIRADS 1-2 vs. 3-5 and BIRADS 1-3 vs. 4-5 are much more difficult for single-output networks, with the proposed model achieving almost double TNR at TPR of 0.95. Direct prediction of BIRADS labels achieves the worst results on all tasks, which is also not surprising, considering rate of disagreement between the radiologists on this task ([10]) 

\section{Discussion}
\label{discussion}

We presented a way to derive multiple tasks from any binary or multi-class malignancy classification problem, by different thresholdings of ordinal labels. The proposed multi-task method improves the performance of single-task classifiers. The most significant improvement is achieved on more difficult tasks that suffer from poor data balancing or noise. It appears that the fact that some of the new sub-tasks can train on better-balanced data splits, improves the overall performance of the multi-task method. In this sense, it resembles the concept of curriculum learning, where a network is trained on progressively more difficult tasks, only here the tasks are presented simultaneously.
While the method shown here was originally developed for medical imaging, it is applicable to all fields that utilize ordinal labels, such as quality assessment and risk evaluation.

\section*{References}

\medskip

\small

[1] Bengio, Y., Louradour, J., Collobert, R. and Weston, J., 2009, June. Curriculum learning. In Proceedings of the 26th annual international conference on machine learning (pp. 41-48). ACM.

[2] Chen, B.C., Chen, L., Ramakrishnan, R. and Musicant, D.R., 2006, April. Learning from aggregate views. In Data Engineering, 2006. ICDE'06. Proceedings of the 22nd International Conference on (pp. 3-3). IEEE.

[3] Geras, K.J., Wolfson, S., Shen, Y., Kim, S., Moy, L. and Cho, K., 2017. High-resolution breast cancer screening with multi-view deep convolutional neural networks. arXiv preprint arXiv:1703.07047.

[4] Gulshan, V., Peng, L., Coram, M., Stumpe, M.C., Wu, D., Narayanaswamy, A., Venugopalan, S., Widner, K., Madams, T., Cuadros, J. and Kim, R., 2016. Development and validation of a deep learning algorithm for detection of diabetic retinopathy in retinal fundus photographs. Jama, 316(22), pp.2402-2410.

[5] He, K., Zhang, X., Ren, S. and Sun, J., 2015. Delving deep into rectifiers: Surpassing human-level performance on imagenet classification. In Proceedings of the IEEE international conference on computer vision (pp. 1026-1034).

[6] Li, X., Kao, Y., Shen, W., Li, X. and Xie, G., 2017, March. Lung nodule malignancy prediction using multi-task convolutional neural network. In Medical Imaging 2017: Computer-Aided Diagnosis (Vol. 10134, p. 1013424). International Society for Optics and Photonics.

[7] Liu, S., Pan, S.J. and Ho, Q., 2017, August. Distributed multi-task relationship learning. In Proceedings of the 23rd ACM SIGKDD International Conference on Knowledge Discovery and Data Mining (pp. 937-946). ACM.

[8] Nishikawa, R.M., Comstock, C.E., Linver, M.N., Newstead, G.M., Sandhir, V. and Schmidt, R.A., 2016, June. Agreement Between Radiologists’ Interpretations of Screening Mammograms. In International Workshop on Digital Mammography (pp. 3-10). Springer, Cham.

[9] Read, J. and Hollmén, J., 2015. Multi-label classification using labels as hidden nodes. arXiv preprint arXiv:1503.09022.

[10] Salazar, A.J., Romero, J.A., Bernal, O.A., Moreno, A.P. and Velasco, S.C., 2017. Reliability of the BI-RADS Final Assessment Categories and Management Recommendations in a Telemammography Context. Journal of the American College of Radiology, 14(5), pp.686-692.

[11] Steinberg, D.M., Fine, J. and Chappell, R., 2008. Sample size for positive and negative predictive value in diagnostic research using case–control designs. Biostatistics, 10(1), pp.94-105.

\end{document}